\pdfoutput=1
\documentclass[conference]{IEEEtran}
\usepackage{cite}
\usepackage{amsmath,amssymb}
\usepackage{graphicx}
\usepackage{booktabs}
\usepackage{tabularx}
\usepackage{tikz}
\usepackage{algorithm}
\usepackage[noend]{algpseudocode}
\usepackage[hidelinks]{hyperref}
\graphicspath{{figures/}}

\begin{document}

\title{HoopMind: A Real-Time Neural Game-Tree System for
Opponent-Aware Possession Planning}

\author{
\IEEEauthorblockN{Yibo Gong}
\IEEEauthorblockA{\textit{Beijing National Day School}\\
\textit{(High School Student)}\\
Beijing, China\\
briangongyibo@outlook.com}
\and
\IEEEauthorblockN{Cong Guo}
\IEEEauthorblockA{\textit{University of Memphis}\\
\ \\
Memphis, TN, USA\\
cguo@memphis.edu}
\and
\IEEEauthorblockN{Jiacheng Ding}
\IEEEauthorblockA{\textit{University of Memphis}\\
\ \\
Memphis, TN, USA\\
jding2@memphis.edu}
}

\maketitle

\begin{abstract}
School coaches prepare for opponents with game film and intuition. The
analytics tools of professional teams stay out of reach. We ask how far
\emph{public} data can close this gap. Professional basketball is our case
study, chosen for its data rather than the league. We fuse five public
sources into one per-shot dataset of 4.23M shots over 21 seasons. The sources
are shot locations, two play-by-play feeds, official matchup tracking, and
player biometrics. Alignment across them is 99.5\% to 100\%. We also report
two data pitfalls that are easy to miss. We then model a half-court
possession as a sequential game. Shot values come from \emph{ShotNet}, an
embedding multilayer perceptron (MLP). On a held-out season it beats a
zone-rate baseline and a logistic baseline, and its probabilities are well
calibrated. A depth-limited expectimax search then solves the offensive
decision tree, with branch-and-bound pruning to keep it real time. All
training runs offline, so the online system stays light. A scouting planner
and a playable simulator both run in a single browser page.
\end{abstract}

\begin{IEEEkeywords}
sports analytics, multi-source data fusion, spatiotemporal data mining,
interactive visualization, serious games
\end{IEEEkeywords}

\section{Introduction}
This project began with something we noticed as student players. A school
coach preparing for an opponent has only game film, a whiteboard, and
intuition. Telling players to open a video game is not instruction, and
purely verbal scouting is hard to absorb. A coach could use something in
between: a fast, visual tool, grounded in data, that shows \emph{where,
against whom, and when} an opponent is weak, and lets players feel the answer
by playing against it.

Such a tool needs historical data at scale, and school basketball has none.
Professional basketball does. Three kinds of data are public there: shot
locations, event-level play-by-play, and, since the 2017-18 season,
matchup-level tracking of who guarded whom~\cite{ds_shots, ds_pbp,
ds_nbadata}. Two gaps remain. First, nobody has fused these sources per shot
in the open. Second, the resulting models must run in real time on a school
laptop.\footnote{Code: \url{github.com/sujo666/hoopmind}; demo:
\url{sujo666.github.io/hoopmind}. Review is single-blind, so the links are
open. Y. Gong, a high school student, is the primary contributor.}

This project differs from earlier course and club work. There the first
author trained standard classifiers on tabular sports data and drew static
shot charts. That work made the gap clear. A model that only scores shots
after the fact cannot tell a player what to do next. That is why we added the
planning layer.

The work has three parts.

The first part is the data (\S\ref{sec:data}). Every shot carries its
pre-shot score margin, its assist source, a clutch flag, the identity of any
blocker, and the shooter's height and weight. Alignment is 99.5\% to 100\%,
and the sources agree on 99.9\% of matched outcomes. We also report two
pitfalls that quietly corrupt a naive merge (\S\ref{sec:pitfalls}).

The second part is the models (\S\ref{sec:models}). We build opponent
profiles at four levels: team by zone, individual defender, quarter of the
game, and the single shot. The shot level is ShotNet, an embedding MLP fit to
3.8M training shots. It beats a zone-rate baseline and a logistic baseline on
a held-out season under a strict time split. Every rate estimate uses
empirical-Bayes shrinkage, and we state where the matchup data is biased.

The third part is planning and the system (\S\ref{sec:game},
\S\ref{sec:system}). We treat a half-court possession as a sequential game
and solve it by depth-limited expectimax with branch-and-bound pruning. Two
browser tools ship the result: a scouting planner that answers where, against
whom, and when to attack, and a playable simulator whose defense follows the
learned profiles. Every shot in the simulator is one in-browser forward pass
of ShotNet, and a coach mode shows the live solution (Fig.~\ref{fig:system}).

\begin{figure}[t]
\centering
\includegraphics[width=0.99\columnwidth]{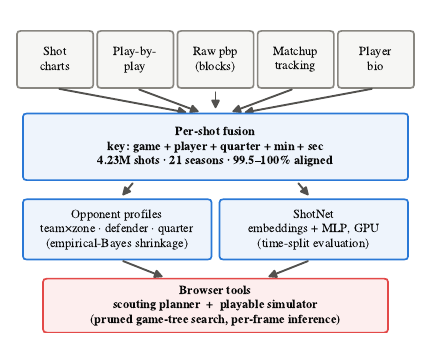}
\caption{System overview: five public sources fused per shot, profiles and
the shot model learned offline, both feeding two browser front-ends.}
\label{fig:pipeline}
\end{figure}

\section{Related Work}
\textbf{Spatial and tracking models.} These works build the two quantities we
need: where shots come from, and what a defense does to them. Reich et
al.~\cite{reich2006} model shot charts spatially, and later work splits shot
intensity into spatial bases~\cite{miller2014}. We use the same zone view of
the court. Franks et al.~\cite{franks2015} estimate defensive skill, Cervone
et al.~\cite{cervone2016} expected possession value, and Le et
al.~\cite{le2017} learn ``ghost'' defenders by imitation. Those three are the
closest in goal to our planner, but all three need private optical tracking.
We ask how far the same goals go on public data alone.

\textbf{Shot outcome prediction.} This line sets the bar ShotNet must clear.
Models built from public features report an area under the ROC curve (AUC)
of about 0.60 to 0.68~\cite{zuccolotto2018}. Adding the closest-defender
distance raises accuracy, but that feature is private. ShotNet sits inside
the public band, so our focus is the fused features, the time-based
evaluation, and the interactive deployment, not a higher AUC.

\textbf{Shot selection and decision value.} This line asks our exact
question: not where players shoot, but \emph{whether they should}.
Skinner~\cite{skinner2012} derives optimal shot thresholds analytically. We
use the same expected-points idea, but put it inside a searchable action tree
tied to one named opponent. Seidl et al.~\cite{seidl2018} let a user sketch
plays against a learned defense, the same interactive goal as our coach mode,
again from private tracking.

\textbf{Sports video games.} Our simulator borrows its runtime design from
here. Commercial titles drive non-player characters with finite-state
machines, influence maps, and steering behaviors~\cite{buckland2005}, and we
keep that architecture: perception delay, then steering, then simplified
execution. Their player ratings, however, are set by editors and never
checked against outcomes. We replace them with estimates from data, checked
on a held-out season.

\section{Possession as a Pruned Sequential Game}\label{sec:game}

\subsection{Formulation}
A five-on-five half-court possession is a two-team zero-sum stochastic
(Markov) game~\cite{littman1994}
$\mathcal{G}=(\mathcal{S}, \mathcal{A}_O, \mathcal{A}_D, T, r)$. A state $s\in\mathcal{S}$ holds the ball location, the positions of all ten
players, the distance from each attacker to the nearest defender, both
clocks, and the score margin. The offense picks an action from its playbook
$\mathcal{A}_O$: shoot, drive, pass, screen, cut, and so on. The defense
picks a coverage from $\mathcal{A}_D$. The reward $r$ is the points scored on
the possession. Solving $\mathcal{G}$ exactly in real time is out of reach,
so we make three assumptions:

\textbf{A1 (empirical opponent).} The defense plays a behavioral policy
$\pi_D$ estimated from data rather than a worst-case adversary. We write
$\pi_D=(\rho,\delta,\iota)$ for its three estimated tables, all learned in
\S\ref{sec:models}. Here $\rho[t,z]$ is what team $t$ concedes in zone $z$,
$\delta[i]$ is how attackable defender $i$ is, and $\iota[t,q]$ is how hard
team $t$ defends in quarter $q$. Fixing one side's policy in a Markov game
leaves a Markov decision process (MDP) for the other
side~\cite{littman1994}. Planning is therefore well defined.

\textbf{A2 (where the randomness sits).} Contest and block uncertainty go
into the terminal payoff. That payoff is linear in the outcome
probabilities, so taking their expectation at the leaf loses nothing. The
only explicit chance nodes are turnovers. A drive or a pass succeeds with
probability $1-\lambda_a$. We set $\lambda_{\mathrm{dr}}{=}0.06$ for a drive
and $\lambda_{\mathrm{ps}}{=}0.035$ for a pass, from the turnover rates in
the development seasons. Otherwise the possession ends with zero points.

\begin{figure}[t]
\centering
\begin{tikzpicture}[scale=1.12, transform shape,
  font=\footnotesize,
  st/.style={draw=black!55, rounded corners=1.5pt, fill=blue!7, inner sep=2.2pt},
  tm/.style={draw=black!55, rounded corners=1.5pt, fill=red!8, inner sep=2.2pt},
  ch/.style={draw=black!55, circle, fill=black!8, inner sep=1.3pt},
  e/.style={-stealth, black!60, thin}]
\node[st] (root) at (0,0) {$s$: on-ball state};
\node[tm] (shoot) at (-2.45,-1.05) {SHOOT: EP$(s)$};
\node[ch] (cd) at (-0.15,-1.05) {};
\node[ch] (cp) at (1.85,-1.05) {};
\draw[e] (root) -- (shoot);
\draw[e] (root) -- node[pos=.62,left=-1pt,font=\scriptsize]{drive$\times3$} (cd);
\draw[e] (root) -- node[pos=.62,right=-1pt,font=\scriptsize]{pass$\times4$} (cp);
\node[font=\scriptsize, black!60] at (-0.15,-1.52) {$\lambda_{\mathrm{dr}}$};
\node[font=\scriptsize, black!60] at (1.85,-1.52) {$\lambda_{\mathrm{ps}}$};
\node[st] (sd) at (-1.30,-2.05) {$s'$: help comes};
\node[st] (sp) at (2.05,-2.05) {$s''$: ball at $j$};
\draw[e] (cd) -- (sd);
\draw[e] (cp) -- (sp);
\node[tm] (fin) at (-2.72,-3.05) {FINISH};
\node[tm] (kick) at (-0.85,-3.05) {KICK-OUT};
\node[tm] (cs) at (1.15,-3.05) {C\&SHOOT};
\node[st, dashed] (ret) at (3.05,-3.05) {$V(\cdot,d{-}1)$};
\draw[e] (sd) -- (fin); \draw[e] (sd) -- (kick);
\draw[e] (sp) -- (cs); \draw[e] (sp) -- (ret);
\draw[black!45, dash pattern=on 1.6pt off 1.3pt] (2.40,-2.72) -- (2.72,-2.40);
\node[font=\scriptsize, black!55] at (2.06,-2.62) {cut};
\end{tikzpicture}
\caption{The decision (sub)tree the planner expands for one on-ball state. Red
nodes are terminals priced by the learned estimators, circles are turnover
chance nodes ($\lambda$), and the dashed mark is a branch-and-bound cut, taken
when $U(a,s)\le V_{\text{best}}-\varepsilon$. Screens and off-ball cuts enter
through the defender-reaction model rather than as explicit branches.}
\label{fig:tree}
\end{figure}
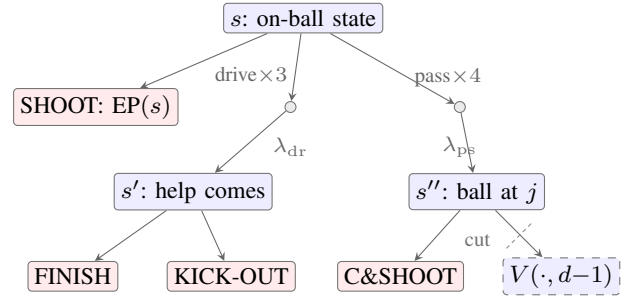

\textbf{A3 (finite horizon).} Search depth is limited to $d$. The search
expands the subset $A(s)\subset\mathcal{A}_O$ shown in Fig.~\ref{fig:tree}.
SHOOT is available in every state. Every leaf of the cut-off tree is
therefore a real terminal, and we never have to guess the value of an
unfinished branch.

Write $\mathrm{EP}(s)$ for the expected points (EP) of shooting from state
$s$. Under A1 to A3 the value of an on-ball state is the depth-limited
expectimax
\begin{equation}
\begin{aligned}
V(s,d)=\max\Big\{\; &\underbrace{\mathrm{EP}(s)}_{\text{shoot now}},\\[-1pt]
&\max_{a\in A(s)}(1-\lambda_a)\, V\big(f(s,a,\pi_D),\, d{-}1\big)\Big\},
\end{aligned}
\label{eq:expectimax}
\end{equation}
The terminal payoff is
$\mathrm{EP}(s) = (1-\beta_{\text{blk}}(s))\cdot p_{\text{make}}(s)\cdot
\mathrm{pts}(s)$. Here $p_{\text{make}}$ is the ShotNet forward pass of
\S\ref{sec:models}, adjusted by the openness and contest layer.
The block probability $\beta_{\text{blk}}$ is read from $\rho$ and $\delta$.
The transition $f(s,a,\pi_D)$ is where the defense responds. A drive moves
the ball 8 to 10 feet toward the rim, tightens the on-ball defender, and
pulls in the weak-side helper. That \emph{frees} every weak-side teammate and
re-prices their shots. A pass moves the ball in $0.9$ seconds, and every
defender recovers toward its mark (Fig.~\ref{fig:tree}).

\subsection{Pruning and complexity}
The action set has $|A(s)|=8$ entries: one shoot, three drives, and four
passes. At depth $d{=}2$ the exhaustive tree evaluates $57.0$ leaf shots on
average. We stop there because deeper search cost more browser time without
changing the action it picked. We prune with branch-and-bound, in the
spirit of alpha-beta search~\cite{knuth1975}. Algorithm~\ref{alg:search}
gives the procedure. Children are ordered by a fast optimistic bound
$U(a,s)$. For a drive that bound is an uncontested rim finish. For a pass it
is a near-open catch-and-shoot at the receiver's spot. A branch is cut when
$U(a,s)\le V_{\text{best}}-\varepsilon$, where $V_{\text{best}}$ is the best
value found so far. At most $k$ children expand per node. Note that $U$ is a
heuristic, not a provably admissible bound, so the pruning is
\emph{approximate}. We therefore measure what it costs instead of claiming
it is exact. 
\begin{algorithm}[t]
\caption{Pruned depth-limited possession search}
\label{alg:search}
\begin{algorithmic}[1]
\Require state $s$, depth $d$, beam $k$, margin $\varepsilon$, defense $\pi_D$
\State $V_{\text{best}} \gets \mathrm{EP}(s)$ \Comment{shoot now: ShotNet + contest layer}
\State $\textit{best} \gets (\textsc{Shoot},\, V_{\text{best}})$
\If{$d = 0$} \Return \textit{best} \EndIf
\State $C \gets \{(a,\, U(a,s)) : a \in \textsc{Drive}_{\{L,M,R\}} \cup \textsc{Pass}_{1..4}\}$
\State sort $C$ by the optimistic bound $U$, descending
\For{$(a, U_a)$ in the first $k$ entries of $C$}
  \If{$U_a \le V_{\text{best}} - \varepsilon$} \textbf{break} \Comment{branch-and-bound cut}
  \EndIf
  \State $s' \gets f(s, a, \pi_D)$ \Comment{defensive response to $a$}
  \State $(\textit{tail}, v') \gets$ \textsc{Search}$(s', d-1, k, \varepsilon)$
  \State $v_a \gets (1-\lambda_a)\, v'$ \Comment{turnover risk on $a$}
  \If{$v_a > V_{\text{best}}$} $V_{\text{best}} \gets v_a$;
     $\textit{best} \gets (a \Vert \textit{tail},\, v_a)$
  \EndIf
\EndFor
\State \Return \textit{best}
\end{algorithmic}
\end{algorithm}

Table~\ref{tab:prune} reports that cost over 300 random states. The shipped
setting is $k{=}6$ and $\varepsilon{=}0.02$. It evaluates 43\% fewer nodes. It
still picks the same top action as exhaustive search in 97\% of states. That
was the best trade-off we found between computation and agreement. All
training happens offline on a GPU. The browser therefore re-solves the tree
every $0.25$ seconds in well under a millisecond.

\begin{table}[t]
\caption{Pruning trade-off vs.\ exhaustive depth-2 expectimax (300 random
states; ``agree'' = same top-level action as exhaustive search).}
\label{tab:prune}
\centering
\begin{tabularx}{\columnwidth}{@{}X rrr@{}}
\toprule
Config & nodes saved & agree & mean EP gap \\
\midrule
$k{=}4,\ \varepsilon{=}0.05$ & 65.8\% & 85.0\% & 0.022 \\
$k{=}5,\ \varepsilon{=}0.02$ & 55.0\% & 91.3\% & 0.008 \\
$k{=}6,\ \varepsilon{=}0.02$ (used) & \textbf{43.4\%} & \textbf{97.0\%} & \textbf{0.001} \\
\bottomrule
\end{tabularx}
\end{table}

\begin{algorithm*}[t]
\caption{From public files to an on-court recommendation: (a)~the offline
build, (b)~one online frame. Symbols follow \S\ref{sec:game}.}
\label{alg:pipeline}
\begin{minipage}[t]{0.485\textwidth}
\footnotesize
\textbf{(a) Offline: public event streams $\rightarrow$ browser payload}
\begin{algorithmic}[1]
\Require seasons $Y$ and the five public sources of Fig.~\ref{fig:pipeline}
\State $\mathcal{D} \gets \emptyset$
\For{$y \in Y$}
  \State $B \gets$ play-by-play$(y)$; recode \texttt{0-0} as missing; fill forward
  \State $J \gets$ shots$(y) \bowtie B$ on (game, player, quarter, min, sec)
  \State $\textit{margin} \gets$ score margin $-$ points of this shot
  \State attach blocker and biometrics; $\mathcal{D} \gets \mathcal{D} \cup J$
\EndFor
\State $\rho[t,z] \gets \textsc{EBShrink}$(field goal \% allowed, block rate), per team-zone
\State $\delta[i] \gets \textsc{EBShrink}$(pts per 100 targeted poss.), per defender
\State $\iota[t,q] \gets \textsc{EBShrink}$(field goal \% allowed), per team-quarter
\State $\theta \gets \textsc{TrainShotNet}(\mathcal{D})$, split by season (train $\le$ 2021-22)
\State \Return payload $(\rho, \delta, \iota, \theta)$, exported as JSON
\end{algorithmic}
\end{minipage}\hfill
\begin{minipage}[t]{0.485\textwidth}
\footnotesize
\textbf{(b) Online: one frame of coupled five-on-five play}
\begin{algorithmic}[1]
\Require state $s$, payload $(\rho,\delta,\iota,\theta)$, step $\Delta t$
\State $h \gets$ off-ball defender nearest the rim \Comment{designated helper}
\For{each defender $i$, marking attacker $m(i)$}
  \State $\hat{x}_i \gets \hat{x}_i + (1-\mathrm{e}^{-\Delta t/\tau_i})(x_{m(i)} - \hat{x}_i)$
    \Comment{lag $\tau_i$ from $\delta[i]$}
  \State $g \gets$ point on the $\hat{x}_i$-to-rim line, tighter if $i$ guards the ball
  \If{ball in the paint \textbf{and} $i{=}h$} $g \gets$ midpoint(ball, rim) \EndIf
  \State steer $i$ to $g$: speed $\propto \iota$ for this quarter, bounded acceleration
\EndFor
\State push apart defenders closer than $2.4$\,ft; each off-ball attacker
  cuts backdoor if its defender left, else re-spaces
\State $o_j \gets$ openness of attacker $j$ from defender distance and momentum
\State $\mathrm{EP}_j \gets \textsc{Price}(\theta, \rho, x_j, o_j, \text{clock}, \text{margin})$
\State \Return \textsc{Search}$(s, 2, k, \varepsilon)$ \Comment{Alg.~\ref{alg:search}, every $0.25$\,s}
\end{algorithmic}
\end{minipage}
\end{algorithm*}

\section{Data and Fusion}\label{sec:data}

\subsection{Sources and alignment}
Algorithm~\ref{alg:pipeline}(a) gives the whole offline path, from the raw
public files to the payload the browser loads. The five sources of
Fig.~\ref{fig:pipeline} differ in coverage. Shot charts run from 2003-04
through 2023-24, while both play-by-play feeds and the
biometrics~\cite{ds_bio} run from 1996-97. Matchup tracking starts in
2017-18, when the league began publishing who guarded whom. We join shots to
play-by-play on the key
\texttt{(game, player, quarter, min-left, sec-left)}. Across 21 seasons the
match rate is 99.5\% to 100\%. The two sources also agree on 99.9\% of matched
outcomes, since both come from the same official scorer feed. Blocks come
from the third-player slot of the raw play-by-play. Court zones are the 14
combinations of range and side labels in the league's application programming
interface (API): the restricted area, three paint sectors, five mid-range
sectors, and five three-point sectors. Backcourt heaves fall outside this
scheme. They are $0.2\%$ of attempts, and we drop them everywhere below.

\subsection{Two pitfalls worth recording}\label{sec:pitfalls}
\textbf{Zero versus missing.} In play-by-play files up to 2016-17, a
non-scoring event carries \texttt{0-0} rather than an empty score. Filling
scores forward then gives margin~0 for most shots. That inflates the rate of
``clutch'' shots, meaning the last 5 minutes with
$|\text{margin}|\le 5$, from about $4\%$ to about $7.5\%$. The give-away is a
jump at the 2017/2018 file boundary.

\begin{figure*}[t]
\centering
\includegraphics[width=0.99\textwidth]{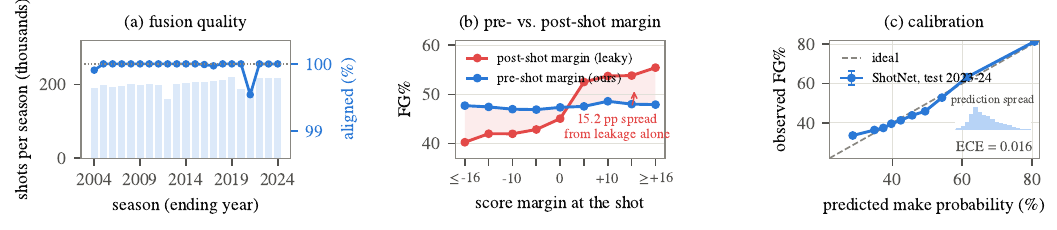}
\caption{(a)~Alignment over 21 seasons: 4.23M shots, never below 99.54\%
matched (bars: volume; line: rate). (b)~The post-shot margin (red) creates a
15.2 percentage point effect that the pre-shot margin (blue) shows to be an
artifact.
(c)~Reliability on the held-out season, with the prediction spread inset.}
\label{fig:eval}
\end{figure*}

\textbf{Post-shot score leakage.} A score is recorded \emph{after} the shot
it accompanies. Used as-is, it produces a strong and steady trend. Teams
ahead by 16 or more appear to shoot 55.4\%, against 40.3\% when behind by 16
or more (Fig.~\ref{fig:eval}(b), red). We then subtract the points of the
shot itself. The curve flattens to a range of 46.9\% to 48.6\% across all
margin bins (blue). The effect was almost entirely an artifact of the
encoding. Every margin in this paper is a pre-shot margin.

\subsection{What fusion unlocks}
\begin{figure}[t]
\centering
\includegraphics[width=0.82\columnwidth]{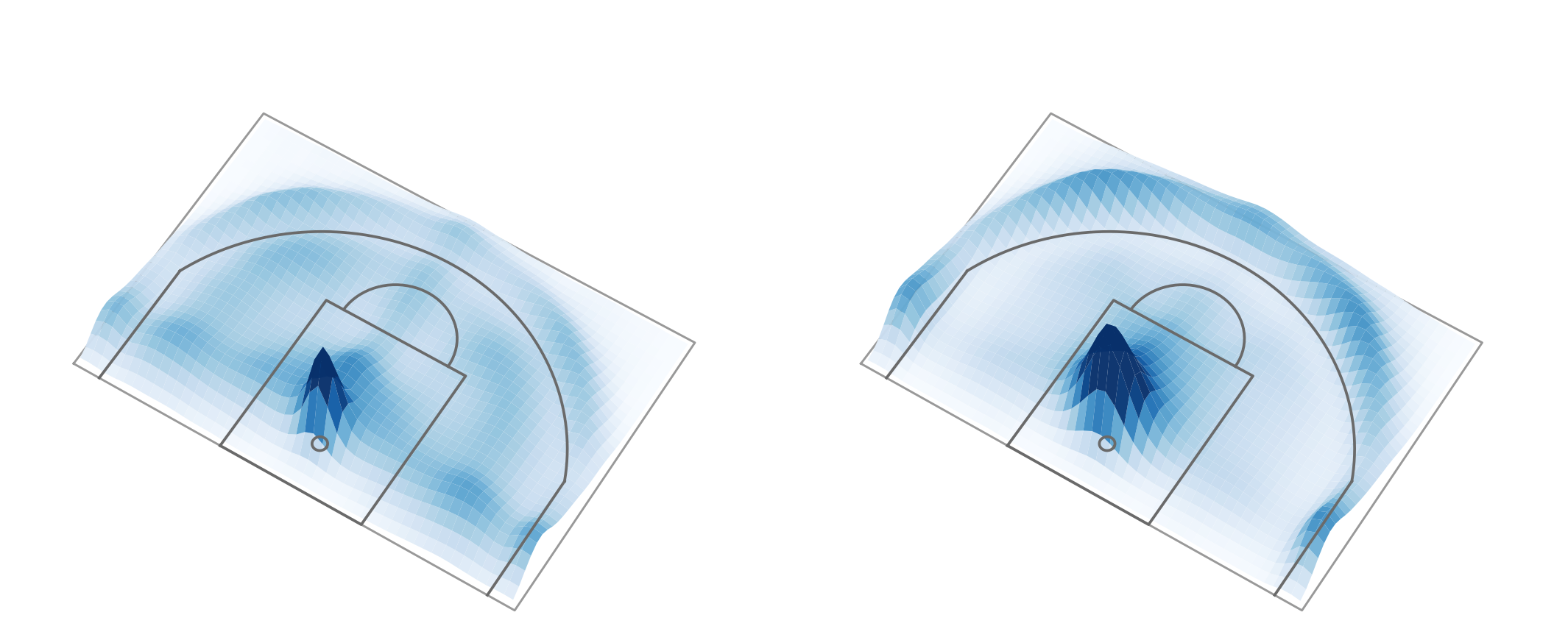}
\caption{Shot-density landscapes, first versus last season (square-root
scale, shared normalization). The mid-range plateau drains into a ridge along
the arc.}
\label{fig:court}
\end{figure}

Here are three findings from 2023-24. First, field goal percentage (FG\%)
falls under pressure. In the clutch, meaning the last five minutes with
$|\text{margin}|\le5$, it drops from 47.7\% to 43.9\%. Three-point (3PT) shots
fall hardest, from 36.9\% to 31.7\%. We can measure this only because the
pre-shot margin was fused in. Second, corner threes are assisted on 96\% to
97\% of makes, against only 46\% for mid-range makes, so court position
decides how a shot is created. Third, 9.0\% of two-point attempts are
blocked, against 0.9\% of threes. Fig.~\ref{fig:court} shows the
structural shift across the 21 seasons. The mid-range emptied out, from
35.6\% of attempts to 11.2\%. The three-point share more than doubled, from
18.7\% to 39.5\%.

\section{Learned Opponent Profiles}\label{sec:models}

\subsection{Team, defender, and time profiles}
For every defending team and every one of the 14 zones we estimate two
numbers: the FG\% it allows and the rate at which it blocks. Both use
empirical-Bayes shrinkage toward the league rate, with a prior strength of
100 attempts. A sanity check supports them. In 2023-24 the best rim defense belongs to
Oklahoma City, Cleveland, and Minnesota, the teams of the rim protectors
Holmgren, Mobley, and Gobert. The weakest belong to Washington, Portland, and
Toronto. From matchup tracking we then rank each roster's defenders. The measure is points allowed per 100 partial possessions
when a defender is targeted, again shrunk, over at least 400 possessions.
Boston is a good example of the spread. Its backup centers concede 34.5
points per 100 targeted possessions. Its reserve guards concede 15.5. The
planner attacks that gap directly. Per-quarter FG\% allowed gives each team an
intensity curve. Boston, for instance, defends hardest in the first quarter,
at $3.1$ percentage points (pp) below its own average, and eases off late.

\textbf{Assignment bias.} Matchup possessions are not assigned at random.
Coaches hide weak defenders on weak scorers. A defender ranking therefore
measures output when targeted, not pure ability. The tools label it that way.
Conditioning on the offensive strength of each defender's assignments would
sharpen the ranking further.

\subsection{ShotNet}
ShotNet learns four embeddings: shooter (16-d), defending team (8-d), zone
(6-d), and action group (6-d). It concatenates them with nine numeric
features: $x$, $y$, distance, a 3PT flag, the quarter, seconds left, the
pre-shot margin, and the shooter's height and weight. A 96-64-1 MLP sits on
top, about 46k parameters in all. It trains in minutes on one workstation
GPU, using AdamW with batch size 16384 for four epochs. Training uses 3.79M
shots from seasons up to 2021-22, and validation uses 217k shots from
2022-23. The 2023-24 test season, 218k shots, is never touched during
development. Table~\ref{tab:results} reports the results.
Fig.~\ref{fig:eval}(c) shows that the model is well calibrated. Its 10-bin
expected calibration error is 0.016, and its mean prediction is 47.4\%
against a 47.5\% base rate. Calibration matters more here than raw AUC,
because Eq.~\eqref{eq:expectimax} consumes probabilities, not rankings. The
margin over the logistic baseline is also not only a metric gap. The planner
needs a prediction conditioned on the shooter and the defending team at the
same time. The embeddings give that, and a numeric-feature model cannot. The
size of the improvement matches prior work on public shooting
features~\cite{zuccolotto2018}, since shot outcomes stay noisy without
defender-distance tracking. The trained network exports to 362\,KB of JSON.
It runs in about $10^4$ multiply-adds, one forward pass per frame in the
browser.

\begin{table}[t]
\caption{Shot-outcome prediction on the held-out 2023-24 season.}
\label{tab:results}
\centering
\begin{tabularx}{\columnwidth}{@{}X rr@{}}
\toprule
Model & AUC & log-loss \\
\midrule
League zone rate & 0.631 & 0.663 \\
Logistic regression (numeric features) & 0.639 & 0.666 \\
\textbf{ShotNet (ours)} & \textbf{0.646} & \textbf{0.648} \\
\bottomrule
\end{tabularx}
\end{table}

\section{The Interactive Layer}\label{sec:system}

\subsection{Scouting planner}
The planner takes one opponent as input. It colors the court by how much
better than league average each zone is to attack. It ranks that roster's
defenders by how attackable they are, and beside each one it shows the
selected player's own history against him. It also shows the per-quarter
intensity strip. For a named attacker, the prediction combines his zone
profile with the opponent's in log-odds space.

\subsection{Playable simulator with a live-solved game tree}
\begin{figure}[t]
\centering
\includegraphics[width=0.93\columnwidth]{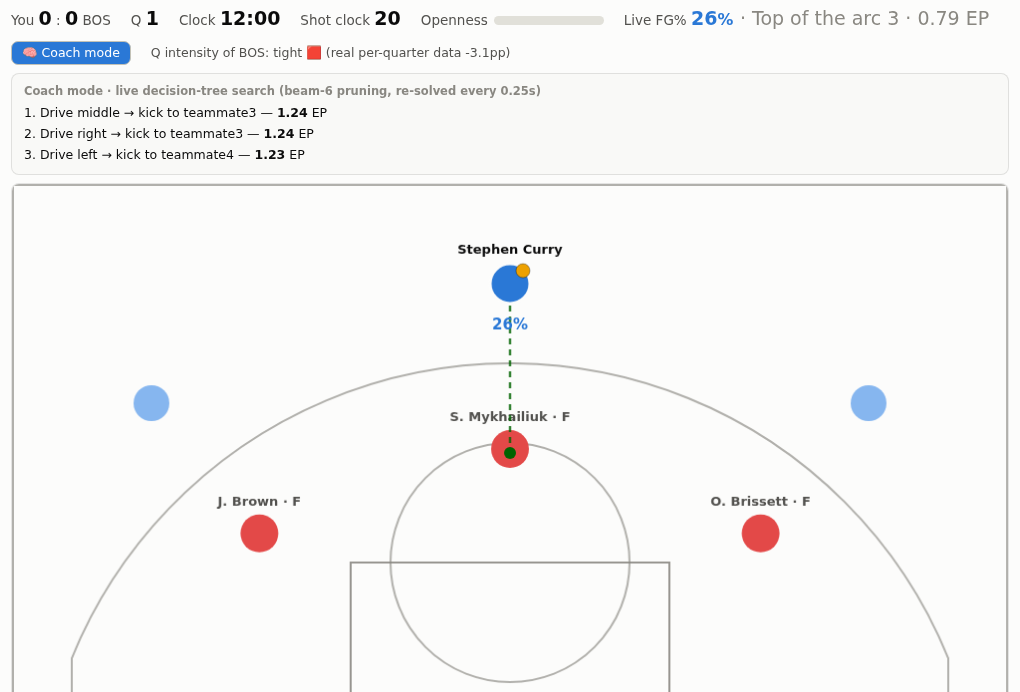}
\caption{The simulator versus Boston's real rotation: live FG\%, coach-mode
top-3 branches with expected points, and the best branch drawn on court.}
\label{fig:system}
\end{figure}

Algorithm~\ref{alg:pipeline}(b) gives one simulation frame. This is where one
action changes the other nine players. When the helper leaves its assignment,
a teammate's shot is re-priced. The search in Algorithm~\ref{alg:search} then
exploits that new price. The simulator keeps the usual sports-game AI stack: perception delay,
steering with bounded acceleration, and man or help
assignment~\cite{buckland2005}. Every parameter, however, comes from the
data. Reaction time comes from the defender's matchup difficulty and ranges
from 0.10 to 0.34 seconds. Speed comes from position group, block attempts
from personal and team-zone block rates, and per-quarter intensity from the
team's own curve. Substitutions at quarter boundaries follow the real
rotation. A shot resolves as one ShotNet forward pass plus a game layer for
openness and contest distance, whose coefficients we show to the player. \emph{Coach mode} (Fig.~\ref{fig:system}) runs
the pruned search of \S\ref{sec:game} every 0.25 seconds. It lists the top
three branches with their expected points and draws the best one on the
court. This is what a scouting analyst would sketch, recomputed as the play
develops. In the state shown, a contested pull-up three against Boston is
worth $0.79$ expected points, while driving middle and kicking to the
weak-side shooter is worth $1.24$. The difference between a habitual shot and
a better branch becomes a number the player can question. A post-game report
lists every shot with its openness, closest defender, model probability, and
outcome, against expected points. One real-time minute compresses to about
one quarter. Each tool is one self-contained HTML file, and a FastAPI backend
serves the same models programmatically.

The interface is explicit about what each number is. Every probability
traces back to a documented estimator in the open pipeline, and defender
\emph{paths} are behavior AI driven by those same estimates, because movement
tracking is not published.

The whole system is reproducible from our public repository. A reader can
clone it and run \texttt{make}. That rebuilds the dataset, the three profile
tables, and ShotNet from the raw public files, and regenerates every number
and figure here. Both browser tools are also hosted live, so a reader can try
them without installing anything.

\subsection{Beyond basketball}
The same framework should carry to other sports with enough public event
data. The recipe has four steps: fuse the public event streams, estimate how
the opponent defends, learn a calibrated pricing model for terminal actions,
then solve the offensive decision tree with pruned search. Any invasion sport
with open event data qualifies. Soccer through StatsBomb's open data and ice
hockey through public shot logs are the two we port to next.

\section{Conclusion}
We combined public shot, play-by-play, matchup, and biometric data into one
per-shot dataset, and used it to estimate opponent profiles and shot
probabilities. Those estimates price the leaves of a depth-limited search,
which runs inside a browser-based simulator. Two results follow. The fusion
holds high alignment across sources. The pruned search cuts computation a lot
and still picks the same action as exhaustive search most of the time.
Basketball is only the case study; the pipeline is what we expect to carry
over.

\bibliographystyle{IEEEtran}
\bibliography{references}

\end{document}